\documentclass[10pt,twocolumn,letterpaper]{article}

\usepackage{cvpr}              

\definecolor{cvprblue}{rgb}{0.21,0.49,0.74}
\usepackage[pagebackref,breaklinks,colorlinks,allcolors=cvprblue]{hyperref}
\usepackage[accsupp]{axessibility}

\usepackage{graphicx}
\usepackage{algorithm}
\usepackage{algpseudocode}
\usepackage{booktabs}
\usepackage{multirow}
\usepackage{color}
\usepackage[table]{xcolor}
\usepackage{listings}
\usepackage{wrapfig}
\usepackage{enumitem}
\usepackage{adjustbox}
\usepackage[most]{tcolorbox}
\usepackage{url}
\usepackage{tablefootnote}
\def\ours{YOLO-NAS-Bench}
\definecolor{Gray}{gray}{0.9}
\definecolor{Blue}{rgb}{0.92, 0.96, 1.0}
\definecolor{LightRed}{rgb}{0.8, 0.3, 0.3}
\definecolor{LightOrange}{rgb}{0.99, 0.92, 0.88}

\def\paperID{} 
\def\confName{CVPR}
\def\confYear{2026}

\title{\ours{}: A Surrogate Benchmark with Self-Evolving Predictors for YOLO Architecture Search}

\author{
Zhe Li \quad 
Xiaoyu Ding \quad 
Jiaxin Zheng  \quad
Yongtao Wang \thanks{Corresponding author.} \\
Wangxuan Institute of Computer Technology, Peking University \\
{\tt\small zheli@stu.pku.edu.cn \quad wyt@pku.edu.cn}
}

\begin{document}
\maketitle
\begin{abstract}
Neural Architecture Search (NAS) for object detection is severely bottlenecked by high evaluation cost, as fully training each candidate YOLO architecture on COCO demands days of GPU time. Meanwhile, existing NAS benchmarks largely target image classification, leaving the detection community without a comparable benchmark for NAS evaluation.
To address this gap, we introduce \textbf{\ours{}}, the first surrogate benchmark tailored to YOLO-style detectors.
\ours{} defines a search space spanning channel width, block depth, and operator type across both backbone and neck, covering the core modules of YOLOv8 through YOLO12.
We sample 1{,}000 architectures via random, stratified, and Latin Hypercube strategies, train them on COCO-mini, and build a LightGBM surrogate predictor.
To sharpen the predictor in the high-performance regime most relevant to NAS, we propose a \emph{Self-Evolving Mechanism} that progressively aligns the predictor's training distribution with the high-performance frontier, by using the predictor itself to discover and evaluate informative architectures in each iteration. This method grows the pool to 1{,}500 architectures and raises the ensemble predictor's $R^2$ from 0.770 to 0.815 and Sparse Kendall Tau from 0.694 to 0.752, demonstrating strong predictive accuracy and ranking consistency.
Using the final predictor as the fitness function for evolutionary search, we discover architectures that surpass all official YOLOv8--YOLO12 baselines at comparable latency on COCO-mini, confirming the predictor's discriminative power for top-performing detection architectures.
The code is available at \url{https://github.com/VDIGPKU/YOLO-NAS-Bench}.
\end{abstract}    
\section{Introduction}
\label{sec:introduction}

\begin{figure}[!t]
  \centering
  \includegraphics[width=1\linewidth]{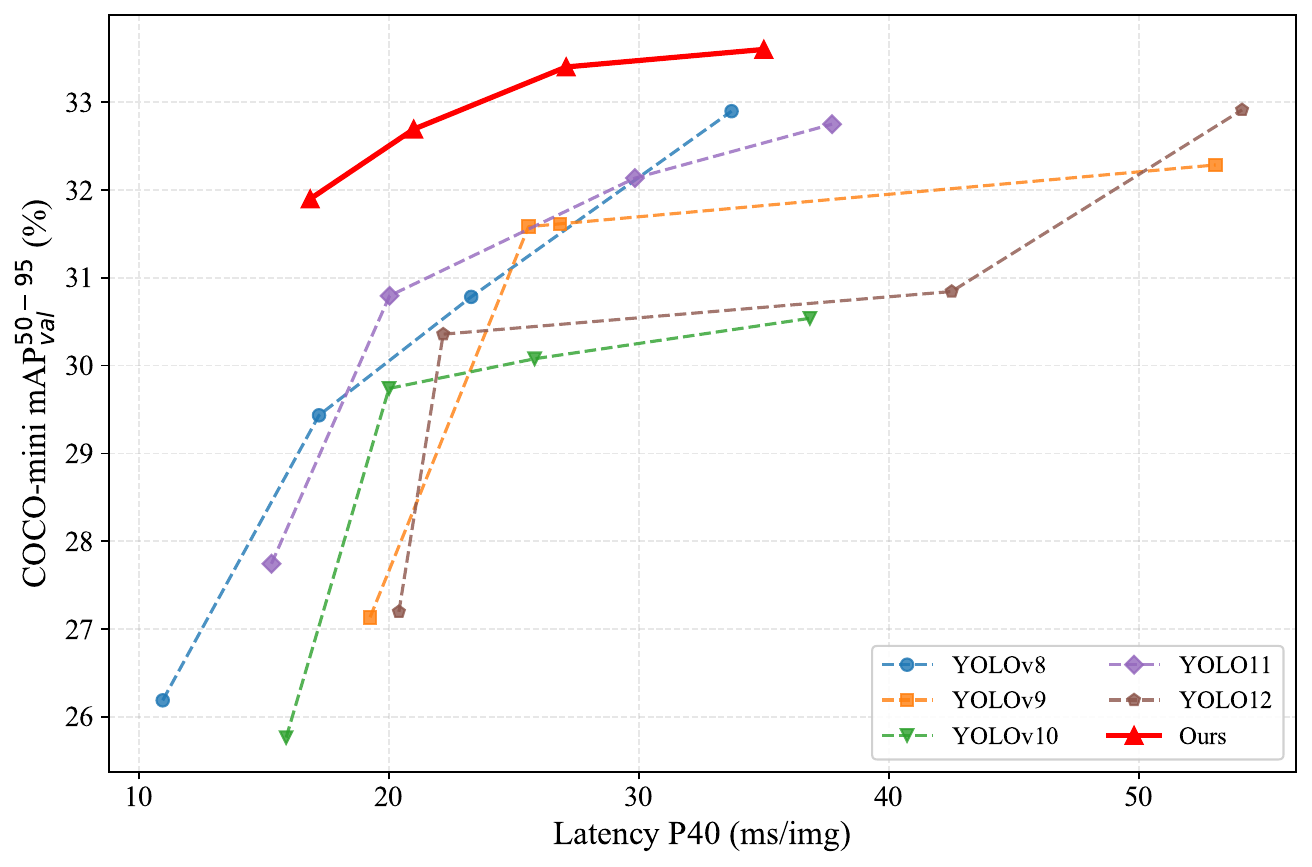}
  \caption{\textbf{Latency vs.\ mAP on COCO-mini.} Architectures discovered by our predictor-guided EA search consistently Pareto-dominate all official YOLO baselines (v8--v12) across the full latency spectrum, demonstrating the strong discriminative power of the \ours{} surrogate predictor.}
  \label{fig:teaser}
\end{figure}

Neural Architecture Search (NAS) demonstrates notable success in automating the design of high-performing image classifiers~\cite{zoph2017neural,liu2018darts,real2019regularized}. However, extending NAS to object detection remains expensive: detection models are much larger, datasets such as COCO~\cite{lin2014coco} require far more compute per training run, and the complete search space needs to jointly consider backbone, neck, and head components. Fully training a single YOLO-family architecture on COCO can take days on a multi-GPU cluster. Evaluating thousands of candidates---as NAS algorithms typically demand---thus becomes infeasible.

NAS benchmarks significantly accelerate research in the classification domain by providing precomputed look-up tables or surrogate predictors, thereby decoupling NAS algorithm development from the expensive architecture evaluation loop and offering a unified benchmark for fair comparison. 
NAS-Bench-101~\cite{ying2019bench} and NAS-Bench-201~\cite{dong2020nasbench201} offer tabular benchmarks with exhaustive evaluation of small cell-based search spaces. NAS-Bench-301~\cite{siems2020bench} scales to the full DARTS space via a surrogate predictor. Yet all these benchmarks target image classification---their search spaces, training pipelines, and evaluation protocols cannot be directly transferred to detection architectures. Meanwhile, detection-specific NAS methods such as Det-NAS~\cite{chen2019detnas}, OPANAS~\cite{liang2021opanas}, and YOLO-NAS~\cite{supergradients} each define bespoke search spaces and evaluation setups, making fair cross-method comparison difficult. The community still lacks unified benchmarks for object detection NAS.

We address this gap with \textbf{\ours{}}, the first surrogate benchmark designed specifically for YOLO-style object detectors. \ours{} features three tightly integrated components: (1)~a comprehensive search space spanning channel width, block depth, and operator type across both backbone and neck, covering the core building blocks of YOLOv8 through YOLO12; (2)~an initial ground-truth database of 1{,}000 architectures sampled via complementary strategies and fully trained on COCO-mini under a unified protocol; and (3)~a \emph{Self-Evolving Predictor} that iteratively enriches the training pool with high-value architectures discovered by evolutionary search, culminating in an ensemble of 10 LightGBM models that achieves sKT\,=\,0.752 and $R^2$\,=\,0.815. We validate the predictor's practical utility by using its predicted mAP as the fitness function for EA-based search~\cite{real2019regularized}. As shown in \cref{fig:teaser}, the discovered architectures \emph{surpass all official YOLO baselines} (v8--v12) at comparable latency.

Our contributions are summarized as follows:
\begin{itemize}[nosep,leftmargin=*]
  \item We design a YOLO-oriented search space covering backbone and neck with channel, depth, and operator dimensions that span the key modules of YOLOv8--YOLO12. Through random, stratified, and Latin Hypercube sampling, we build an initial database of 1{,}000 architectures fully trained on COCO-mini.
  \item We propose a Self-Evolving Predictor that bridges the distribution gap between uniformly sampled training data and the high-performance frontier critical to NAS. The predictor guides latency-bucketed evolutionary search to discover promising architectures, which are trained and fed back to retrain the predictor, forming a self-evolving loop. Over 10 rounds, the pool grows to 1{,}500 architectures and sKT rises from 0.694 to 0.752.
  \item We demonstrate the predictor's practical utility by deploying it as the fitness function for EA search and discovering architectures that surpass official YOLO baselines (v8--v12) at comparable latency on COCO-mini.
\end{itemize}

\section{Related Work}
\label{sec:related-work}

\paragraph{Real-Time Object Detectors.}
The YOLO family has driven rapid progress in real-time detection.
YOLOv8~\cite{yolov8_ultralytics} introduces the C2f module with streamlined cross-stage feature fusion.
YOLOv9~\cite{wang2024yolov9} proposes GELAN and Programmable Gradient Information (PGI) for improved gradient flow.
YOLOv10~\cite{wang2024yolov10} eliminates the NMS post-processing step via consistent dual assignments.
YOLO11~\cite{yolo11_ultralytics} advances efficiency with C3k2 re-parameterized blocks and C2PSA attention modules.
YOLO12~\cite{tian2025yolov12} adopts an attention-centric design, pushing the accuracy--speed Pareto frontier further.
Beyond the YOLO lineage, RT-DETR~\cite{zhao2024detrs} demonstrates that Transformer-based detectors can also achieve real-time speeds.
Despite these continuous improvements, architecture design in the detection domain remains predominantly manual, motivating the need for automated search and, consequently, a standardized benchmark to evaluate NAS algorithms fairly.

\paragraph{NAS Benchmarks.}
Tabular benchmarks such as NAS-Bench-101~\cite{ying2019bench} (423\,k architectures on CIFAR-10) and NAS-Bench-201~\cite{dong2020nasbench201} (15{,}625 cell-based architectures across three datasets) are instrumental in democratizing NAS research for image classification by removing the evaluation bottleneck.
NAS-Bench-301~\cite{siems2020bench} scales to the full DARTS search space by training a surrogate predictor evaluated via the Sparse Kendall Tau (sKT) metric, establishing a protocol that subsequent benchmarks adopt.
NAS-Bench-360~\cite{tu2022bench360} broadens the scope to diverse tasks and data modalities, yet does not include detection-specific search spaces.
YOLOBench~\cite{yang2024yolobench} characterizes a fixed set of YOLO model families but cannot be adapted for NAS algorithm evaluation.
Existing benchmarks predominantly target classification or evaluate fixed model families. Consequently, surrogate benchmarks for detection-oriented NAS remain scarce, and our work addresses this gap.

\paragraph{NAS for Object Detection.}
Several works apply NAS to detection sub-problems:
NAS-FPN~\cite{ghiasi2019nasfpn} searches for feature pyramid topologies;
Det-NAS~\cite{chen2019detnas} performs backbone architecture search with a one-shot supernet;
OPANAS~\cite{liang2021opanas} optimizes FPN neck architectures via an accuracy predictor;
SP-NAS~\cite{jiang2020sp} proposes a serial-to-parallel search strategy for detection backbones;
and YOLO-NAS~\cite{supergradients} employs the proprietary AutoNAC framework to design YOLO architectures.
Each of these methods defines its own search space, training recipe, and evaluation setup, making direct cross-method comparison difficult.
Our paper addresses this fragmentation by providing a shared search space, a precomputed architecture--performance database, and a calibrated surrogate predictor that any NAS algorithm can query at near-zero cost.

\section{Method}
\label{sec:method}

\begin{figure*}[!t]
  \centering
  \includegraphics[width=1\linewidth]{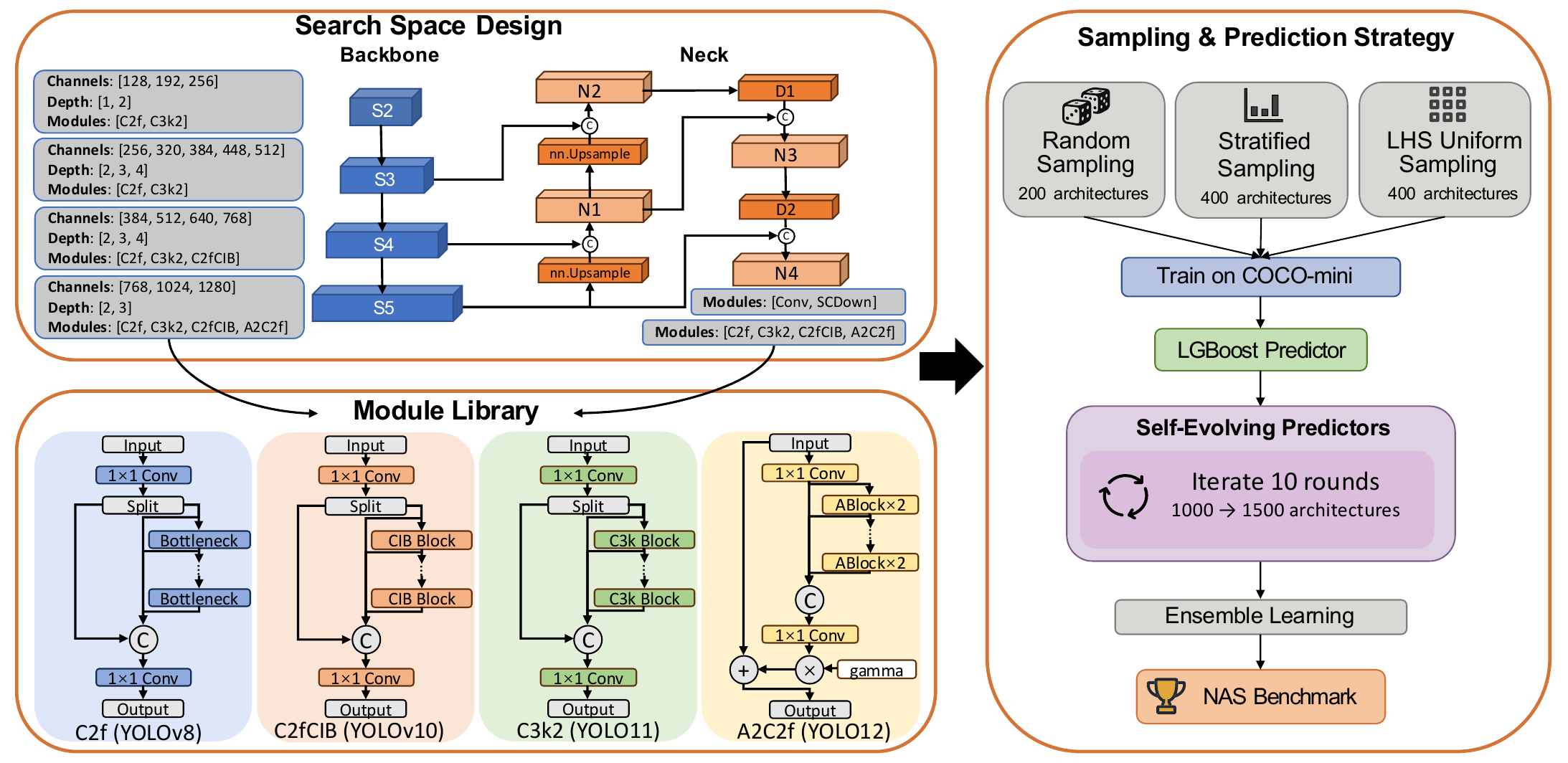}
  \caption{\textbf{Overview of \ours{}.} (1)~A YOLO-style search space spanning channel, depth, and operator dimensions across both backbone and neck is defined. (2)~1{,}000 architectures are sampled via three complementary strategies and trained on COCO-mini. (3)~A LightGBM predictor is trained on the resulting \{architecture, mAP\} pairs. (4)~The Self-Evolving Predictor iteratively expands the pool with high-value architectures discovered by evolutionary search, and retrains the predictor over 10 rounds, yielding an ensemble of 10 LightGBM models over 1{,}500 architectures.}
  \label{fig:pipeline}
\end{figure*}

An overview of \ours{} is illustrated in \cref{fig:pipeline}. Starting from a carefully designed search space over YOLO-style architectures, we sample and fully train 1{,}000 architectures on COCO-mini to build a ground-truth performance database. Each configuration is converted into a compact feature vector, upon which a LightGBM surrogate predictor is trained. 
To further strengthen the predictor in the high-performance regime that matters most for NAS, we propose a \emph{Self-Evolving Predictor} loop that iteratively discovers, trains, and assimilates promising architectures.

\subsection{Benchmark Construction}
\label{sec:method:benchmark-construction}
\paragraph{Search Space Design.}
\label{sec:method:search-space}
We design a comprehensive search space that covers both the backbone and the neck of YOLO-style detectors, while keeping the detection head fixed. The space is parameterized along three dimensions:

\begin{itemize}
    \item \textbf{Channel width:} Each of the four backbone stages (S2--S5) has an independently selectable channel count. The candidate sets grow with the stage depth to reflect the natural widening of feature hierarchies in modern detectors.
    \item \textbf{Block depth:} The number of repeated blocks within each stage is searchable, controlling the representation capacity at each resolution level.
    \item \textbf{Operator type:} This dimension subsumes both the \emph{feature-extraction} module in each stage and the \emph{downsampling} operator between stages. Feature-extraction candidates include C2f~\cite{yolov8_ultralytics}, C3k2~\cite{yolo11_ultralytics}, C2fCIB~\cite{wang2024yolov10}, and A2C2f~\cite{tian2025yolov12}---the core building blocks employed across YOLOv8 through YOLO12---spanning a spectrum from lightweight convolutions (C2f) to efficient re-parameterized blocks (C3k2, C2fCIB) and attention-augmented modules (A2C2f). Downsampling candidates include standard Conv and the stride-channel-decoupled SCDown~\cite{wang2024yolov10}. 
\end{itemize}

In the backbone, the operator palette widens progressively: S2/S3 stages choose between \{C2f, C3k2\}, S4 adds C2fCIB, and S5 further includes A2C2f, reflecting the increasing complexity budget at lower resolutions. 
In the neck, channel width and depth are fixed to avoid excessive growth of the search space; only the operator type is searchable. Feature blocks N1--N4 share a single global choice from \{C2f, C3k2, C2fCIB, A2C2f\}, and the two downsample blocks D1--D2 share a single choice from \{Conv, SCDown\}. Upsample layers use fixed \texttt{nn.Upsample} throughout the YOLO family and are not searchable.

\cref{tab:search-space} shows the full search space specification. The combinatorial product of all dimensions yields an architecture space on the order of millions of unique configurations, providing a rich landscape for NAS algorithm evaluation.

\begin{table}[!t]
  \centering
  \caption{\textbf{Search space of \ours{}.} Channel, depth, and operator choices for each stage.}
  \label{tab:search-space}
  \small
  \begin{tabular}{@{}llll@{}}
    \toprule
    \textbf{Dimension} & \textbf{Stage} & \textbf{Candidates} & \textbf{\#} \\
    \midrule
    \multirow{4}{*}{Channel}
      & S2 & 128, 192, 256 & 3 \\
      & S3 & 256, 320, 384, 448, 512 & 5 \\
      & S4\tablefootnote{S4 channels must be at least S3 to avoid degenerate configurations.} & 384, 512, 640, 768 & 4 \\
      & S5 & 768, 1024, 1280 & 3 \\
    \midrule
    \multirow{4}{*}{Depth}
      & S2 & 1, 2 & 2 \\
      & S3 & 2, 3, 4 & 3 \\
      & S4 & 2, 3, 4 & 3 \\
      & S5 & 2, 3 & 2 \\
    \midrule
    \multirow{3}{*}{\shortstack[l]{Operator\\(backbone)}}
      & S2/S3 & C2f, C3k2 & 2 \\
      & S4 & C2f, C3k2, C2fCIB & 3 \\
      & S5 & C2f, C3k2, C2fCIB, A2C2f & 4 \\
    \midrule
    \multirow{2}{*}{\shortstack[l]{Operator\\(neck)}}
      & N1--N4 & C2f, C3k2, C2fCIB, A2C2f & 4 \\
      & D1--D2 & Conv, SCDown & 2 \\
    \bottomrule
  \end{tabular}
\end{table}

\paragraph{Architecture Sampling.}
\label{sec:method:sampling}
To build a diverse and representative ground-truth database, we employ three complementary sampling strategies over the search space:
(i)~\emph{Random sampling} (200 architectures) provides a uniform baseline coverage;
(ii)~\emph{Stratified sampling} (400 architectures) bins candidates by parameters and draws uniformly within each stratum, ensuring balanced representation across model scales;
(iii)~\emph{Latin Hypercube Sampling}~\cite{mckay1979lhs} (400 architectures) maximizes coverage across the high-dimensional discrete space by applying stratified sampling independently along each search dimension.
The three strategies are complementary: random sampling covers the space broadly, stratified sampling prevents under-representation of light or heavy architectures, and LHS provides near-optimal space-filling properties.

All 1{,}000 sampled architectures are trained from scratch on COCO-mini under an identical training protocol (detailed in \cref{sec:experiment:settings}). COCO-mini is a class- and size-stratified 10\% subset of COCO that preserves the original category and bounding-box size distributions. The resulting database constitutes the foundation of \ours{}.

\paragraph{Architecture Encoding and Surrogate Predictor.}
\label{sec:method:predictor}
Each architecture configuration is converted into a 25-dimensional feature vector. Channel widths and block depths are represented as scalar values, while operator choices are one-hot encoded, allowing the tree-based predictor to split on each operator independently without imposing artificial ordering.

We train a LightGBM~\cite{ke2017lightgbm} gradient-boosted decision tree as the surrogate predictor, regressing from the 25-dimensional encoding to mAP$_{50\text{-}95}$. Following NAS-Bench-301~\cite{siems2020bench}, we adopt two evaluation metrics:
\begin{itemize}[nosep,leftmargin=*]
  \item \textbf{Coefficient of determination} ($R^2$): measures overall regression quality.
  \item \textbf{Sparse Kendall Tau} (sKT): the Kendall $\tau$ rank correlation computed after rounding predictions to $0.1\%$ precision, which discounts ranking changes due to negligible prediction noise. Formally, $\mathrm{sKT} = \tau\!\bigl(\mathbf{y},\, \lfloor \hat{\mathbf{y}} \rceil_{0.001}\bigr)$.
\end{itemize}

\begin{figure}[t]
  \centering
  \includegraphics[width=0.9\linewidth]{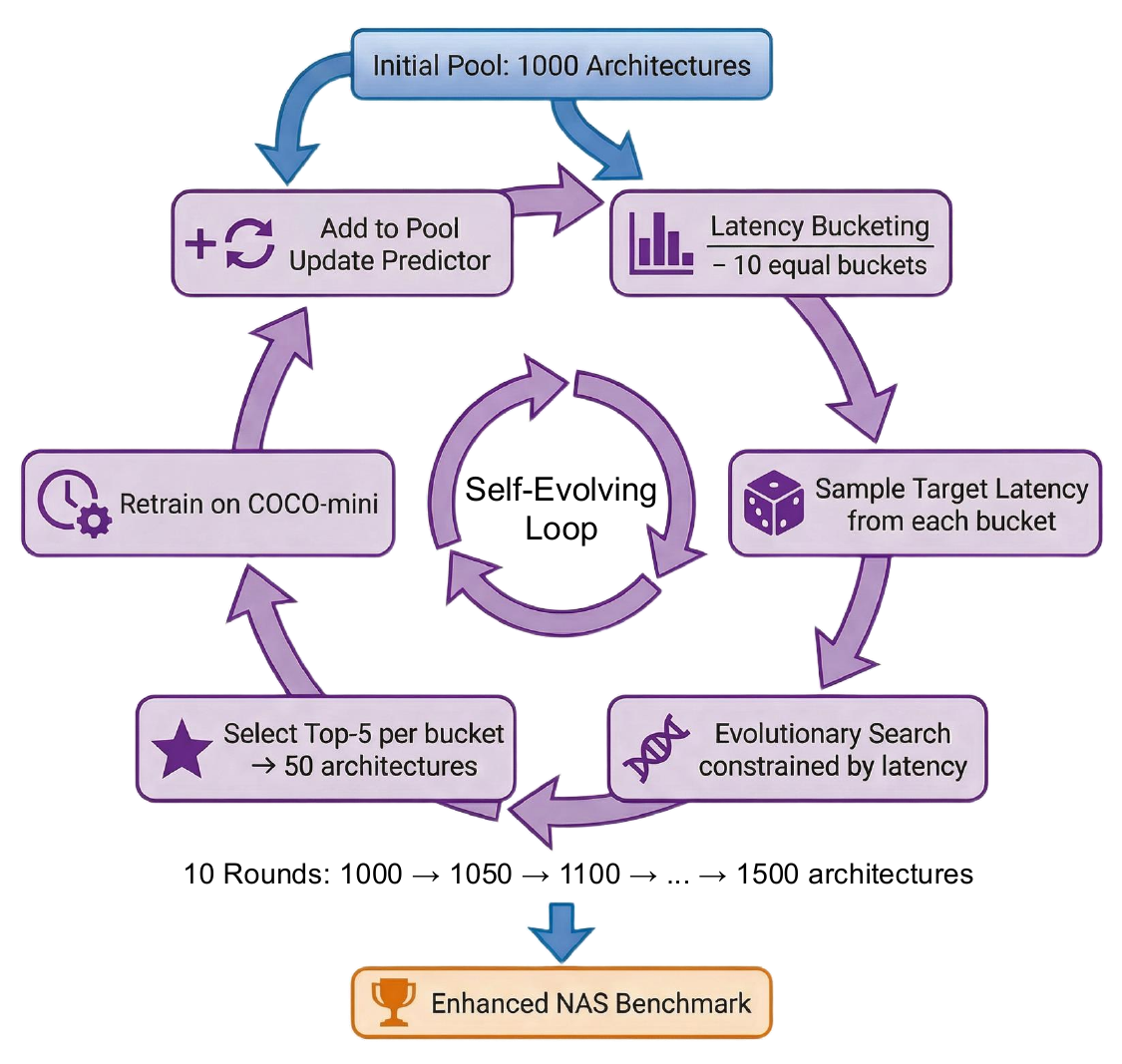}
  \caption{\textbf{Self-Evolving Predictor.} Starting from 1{,}000 architectures, the loop partitions latency into 10 buckets. For each bucket, EA search selects the top 5 architectures using predicted mAP as fitness and real latency as constraint. In each round, these 50 new architectures are trained on COCO-mini, merged into the pool, and the predictor is retrained. After 10 rounds the pool grows to 1{,}500 architectures, yielding an enhanced benchmark enriched in the high-performance regime most relevant to NAS.}
  \label{fig:self-evolving}
\end{figure}

\subsection{Self-Evolving Predictor}
\label{sec:method:self-evolving}
A predictor trained on uniformly sampled architectures may under-represent the high-performance frontier, where ranking accuracy is most critical for NAS. To address this distribution mismatch, we propose a \emph{Self-Evolving Mechanism}. As shown in \cref{fig:self-evolving}, it iteratively enriches the training pool with architectures that the current predictor considers promising, thereby sharpening its discrimination in the region that matters.

\noindent\textbf{Latency bucketing.}
The latency range observed across all 1{,}000 initial architectures is evenly partitioned into 10 buckets. This stratification ensures that the self-evolving loop discovers high-performing architectures at every latency operating point, not only at a single scale.

\noindent\textbf{Evolutionary expansion (single round).}
Within each latency bucket, a target latency is sampled uniformly. An evolutionary algorithm (EA)~\cite{real2019regularized} is then run with predicted mAP as the fitness function and real measured latency as the constraint. 
The EA runs for 100 generations with a population of 50, selecting the top 25\% as parents, generating offspring via crossover (50\%) or mutation (50\%, rate 0.2), and carrying forward the top 10\% as elite. 
The final top 5 architectures from each bucket are selected as candidates, yielding 50 new architectures per round.

\noindent\textbf{Iterative refinement.}
Each batch of 50 new architectures is fully trained on COCO-mini under the same protocol, their configuration and ground-truth mAP are recorded, and the pairs are merged into the architecture pool. The predictor is then retrained on the expanded pool. This loop repeats for 10 rounds, growing the pool from 1{,}000 to 1{,}500 architectures. Crucially, each round's newly added architectures are biased toward high-performance regions identified by the current predictor, progressively narrowing the distribution gap between predictor training and actual NAS search.

\noindent\textbf{Ensemble prediction.}
After the final round, 10 LightGBM models are trained on the full 1{,}500-architecture pool with different random seeds. The ensemble prediction is the arithmetic mean of the 10 models, which reduces variance and further stabilizes ranking quality.

\section{Experiments}
\label{sec:experiment}

\subsection{Experimental Setup}
\label{sec:experiment:settings}

\begin{table}[t]
  \centering
  \caption{\textbf{Unified training configuration} for all architectures in \ours{}.}
  \label{tab:train-config}
  \small
  \begin{tabular}{@{}ll@{}}
    \toprule
    \textbf{Parameter} & \textbf{Value} \\
    \midrule
    Epochs       & 120 \\
    Batch size   & 128 \\
    Image size   & $640 \times 640$ \\
    Learning rate ($\text{lr}_0$) & 0.01 \\
    LR schedule  & Step (decay at epoch 100) \\
    Mosaic       & 1.0 \\
    MixUp        & 0.15 \\
    Copy-Paste   & 0.5 \\
    Pretrained   & No \\
    Workers      & 16 \\
    \bottomrule
  \end{tabular}
\end{table}

\noindent\textbf{Dataset.}
We construct COCO-mini by stratified sampling 10\% of COCO~\cite{lin2014coco} images, preserving the original category distribution and bounding-box size ratios. The resulting subset retains 80 classes and follows the standard train2017/val2017 split. All architectures---both the benchmark pool and YOLO baselines---are trained and evaluated on this identical dataset, ensuring fair comparison.

\noindent\textbf{Training protocol.}
Every architecture is trained from scratch under a unified configuration summarized in \cref{tab:train-config}, including Mosaic~\cite{bochkovskiy2020yolov4}, MixUp~\cite{zhang2018mixup}, and Copy-Paste~\cite{ghiasi2021copy} augmentation. The final mAP$_{50\text{-}95}$ is taken from the last epoch.

\noindent\textbf{Latency measurement.}
All latencies are measured on a single NVIDIA P40 GPU with batch size~1, $640{\times}640$ FP32 input, 10 warmup and 50 timed forward passes. The mean inference time is reported in milliseconds.

\noindent\textbf{Predictor training.}
The surrogate predictor is trained with RMSE as the loss function. We report $R^2$ and sKT on a held-out 20\% validation split.

\subsection{Main Results}
\label{sec:experiment:main-results}

\paragraph{Predictor Quality.}
\label{sec:experiment:predictor}

\begin{figure}[t]
  \centering
  \includegraphics[width=0.73\linewidth]{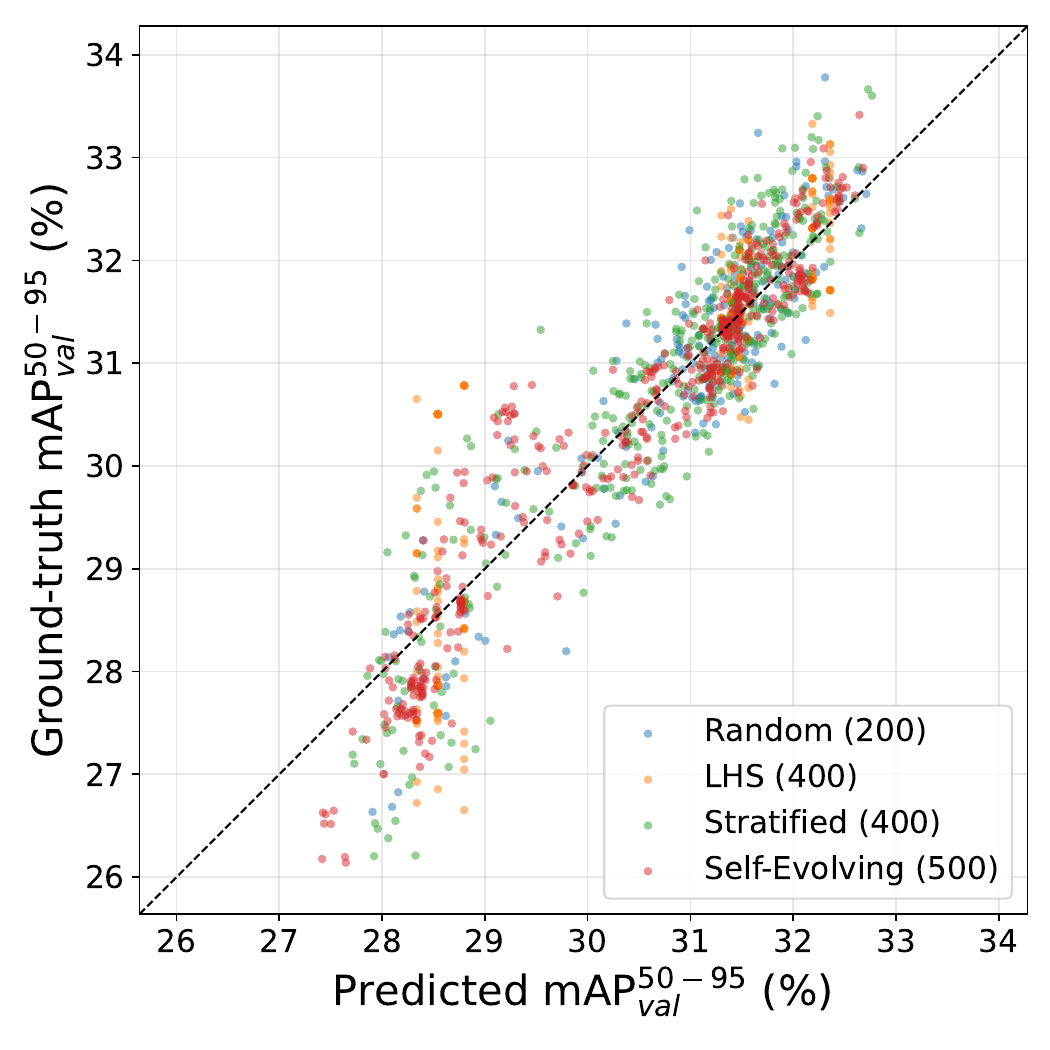}
  \vspace{-0.5em}
  \caption{\textbf{Predicted vs.\ ground-truth mAP on the full 1{,}500-architecture pool.} Each point is an architecture colored by its sampling source. Points cluster closely around the $y{=}x$ diagonal, confirming strong agreement between the ensemble predictor and ground-truth performance.}
  \label{fig:scatter}
\end{figure}

\cref{tab:predictor-quality} summarizes the surrogate predictor's accuracy before and after Self-Evolving. The 10-model LightGBM ensemble trained on the initial 1{,}000 architectures achieves a validation sKT of 0.694. After 10 rounds of Self-Evolving, both $R^2$ and sKT improve: $R^2$ rises from 0.770 to 0.815 (+4.5\%), and sKT from 0.694 to 0.752 (+5.8\%). An sKT of 0.752 indicates strong ranking consistency, while the $R^2$ of 0.815 suggests that the predictor captures the majority of variance in architecture performance. These results show that our surrogate predictor is a high-fidelity proxy for the true performance landscape, and confirm that the Self-Evolving mechanism effectively sharpens the predictor's ranking ability. \cref{fig:scatter} further visualizes the close agreement between predicted and actual mAP across all 1{,}500 architectures.

\begin{table}[t]
  \centering
  \caption{\textbf{Predictor quality before and after Self-Evolving.} Metrics are reported on the validation split (20\%). Ensemble: 10 homogeneous LightGBM models.}
  \label{tab:predictor-quality}
  \small
  \begin{tabular}{@{}lccc@{}}
    \toprule
    \textbf{Setting} & \textbf{\#Archs} & $\boldsymbol{R^2}$ & \textbf{sKT} \\
    \midrule
    Before Self-Evolving & 1{,}000 & 0.770 & 0.694 \\
    After Self-Evolving  & 1{,}500 & \textbf{0.815} & \textbf{0.752} \\
    \bottomrule
  \end{tabular}
\end{table}

\begin{table*}[!t]
  \centering
  \caption{\textbf{Predictor-guided EA search results vs.\ official YOLO baselines} on COCO-mini. All models are trained from scratch under the same protocol. Latency is measured on a single P40 GPU. The top block shows architectures discovered by our predictor-guided EA; subsequent blocks show official baselines.}
  \label{tab:main-results}
  \small
  \setlength{\tabcolsep}{5pt}
  \begin{tabular}{@{}l cc cc cc cc@{}}
    \toprule
    & \multicolumn{2}{c}{\textbf{Small (S)}} & \multicolumn{2}{c}{\textbf{Medium (M)}} & \multicolumn{2}{c}{\textbf{Large (L)}} & \multicolumn{2}{c}{\textbf{Extra-Large (X)}} \\
    \cmidrule(lr){2-3}\cmidrule(lr){4-5}\cmidrule(lr){6-7}\cmidrule(lr){8-9}
    \textbf{Model} & mAP\,(\%) & Lat.\,(ms) & mAP\,(\%) & Lat.\,(ms) & mAP\,(\%) & Lat.\,(ms) & mAP\,(\%) & Lat.\,(ms) \\
    \midrule
    \rowcolor{LightOrange}
    \textbf{Ours} & \textbf{31.9} & 16.85 & \textbf{32.7} & 21.00 & \textbf{33.4} & 27.09 & \textbf{33.6} & 35.00 \\
    \midrule
    YOLO12  & 27.2 & 20.40 & 30.4 & 22.17 & 30.8 & 42.51 & 32.9 & 54.13 \\
    YOLO11  & 27.7 & 15.30 & 30.8 & 20.03 & 32.1 & 29.84 & 32.8 & 37.73 \\
    YOLOv10 & 25.8 & 15.89 & 29.7 & 20.01 & 30.1 & 25.83 & 30.5 & 36.84 \\
    YOLOv9  & 27.1 & 19.25 & 31.6 & 25.59 & 31.6 & 26.85 & 32.3 & 53.07 \\
    YOLOv8  & 26.2 & 10.95 & 29.4 & 17.21 & 30.8 & 23.30 & 32.9 & 33.71 \\
    \bottomrule
  \end{tabular}
  \vspace{-0.5em}
\end{table*}
\paragraph{Predictor-Guided Search Results.}
\label{sec:experiment:main}

To validate that our surrogate predictor can reliably identify top-performing architectures, we conduct an evolutionary architecture search (EA)~\cite{real2019regularized} using the \emph{ensemble predictor's predicted mAP as the fitness function} and real measured latency as the constraint. The EA searches within the same search space defined in \cref{sec:method:search-space}. Top candidate architectures are then \emph{fully retrained} on COCO-mini under the identical protocol of \cref{tab:train-config}, producing ground-truth mAP values.

As shown in \cref{tab:main-results}, the four predictor-discovered architectures Pareto-dominate all official YOLO baselines (v8--v12) across the full latency spectrum, achieving consistently higher mAP at equal or lower latency. The advantage is most pronounced in the small-model regime, where \texttt{Ours-s} surpasses YOLO11s by +4.2\% mAP at comparable latency, and remains substantial at the large end, where \texttt{Ours-x} exceeds YOLO12x in mAP while being $1.5{\times}$ faster.
These results confirm that the predictor not only achieves high ranking correlation (sKT) on held-out data, but also exhibits \emph{strong discriminative power in the top-performance regime}---architectures it deems promising indeed yield superior performance after full retraining.

\begin{table}[!t]
  \centering
  \caption{\textbf{Predictor type comparison} on initial 1{,}000 architectures (no ensemble, same data split).}
  \label{tab:predictor-ablation}
  \small
  \begin{tabular}{@{}lcc@{}}
    \toprule
    \textbf{Predictor} & $\boldsymbol{R^2}$ & \textbf{sKT} \\
    \midrule
    LightGBM  & \textbf{0.768} & 0.699 \\
    XGBoost   & 0.758 & 0.696 \\
    NGBoost   & 0.755 & \textbf{0.704} \\
    Random Forest & 0.744 & 0.678 \\
    MLP       & 0.053 & 0.440 \\
    \bottomrule
  \end{tabular}
\end{table}

\subsection{Ablation Studies}
\label{sec:experiment:ablation}

\paragraph{Predictor type comparison.}
\cref{tab:predictor-ablation} compares five predictor types trained on the initial 1{,}000-architecture pool: LightGBM~\cite{ke2017lightgbm}, XGBoost~\cite{chen2016xgboost}, NGBoost~\cite{duan2020ngboost}, Random Forest~\cite{breiman2001random}, and MLP. Among the tree-based methods, LightGBM achieves the best balance of $R^2$ (0.768) and sKT (0.699), followed closely by NGBoost (sKT 0.704) and XGBoost ($R^2$ 0.758). Random Forest lags slightly behind. The MLP baseline performs poorly. These results justify our choice of LightGBM as the base model for the predictor.

\paragraph{Self-Evolving vs.\ random expansion.}
To disentangle whether the predictor improvement stems from the Self-Evolving mechanism or merely from increasing the training pool size, we conduct an ablation: we train 200 randomly sampled architectures on COCO-mini, add them to the initial 1{,}000 architectures, and retrain the ensemble predictor. For fair comparison, we also truncate Self-Evolving when the pool reaches 1{,}200 architectures. As shown in \cref{tab:self-evolving-ablation}, random expansion yields $R^2=0.776$ and sKT$=0.701$, while Self-Evolving achieves $R^2=0.798$ and sKT$=0.738$. The clear gap (+0.022 $R^2$, +0.037 sKT) confirms that the gain is attributable to the targeted enrichment of high-performance architectures by the Self-Evolving loop, rather than to the increase in data size alone.

\begin{table}[t]
  \centering
  \caption{\textbf{Self-Evolving vs.\ random pool expansion.} Same pool size (1{,}200); metrics on validation split (20\%).}
  \label{tab:self-evolving-ablation}
  \small
  \begin{tabular}{@{}lccc@{}}
    \toprule
    \textbf{Expansion strategy} & \textbf{\#Archs} & $\boldsymbol{R^2}$ & \textbf{sKT} \\
    \midrule
    Initial (no expansion) & 1{,}000 & 0.770 & 0.694 \\
    Random +200 & 1{,}200 & 0.776 & 0.701 \\
    Self-Evolving +200 & 1{,}200 & \textbf{0.798} & \textbf{0.738} \\
    \bottomrule
  \end{tabular}
\end{table}

\subsection{Limitations and Future Work}
The current benchmark is built on COCO-mini (10\% of COCO) and measures latency on a single GPU (NVIDIA P40). To ensure the authenticity of latency measurements across different platforms, we do not build a surrogate latency predictor as we do for mAP; thus, when using this benchmark for NAS, one still needs to measure the latency of each architecture empirically. But this cost is acceptable compared to the error that a latency predictor would introduce on unseen hardware. Extending \ours{} to the full COCO dataset, diverse hardware platforms (edge GPUs, mobile NPUs), and additional tasks such as instance segmentation and pose estimation are natural next steps.

\section{Conclusion}
\label{sec:conclusion}

We present \ours{}, the first surrogate benchmark tailored to YOLO-style detectors. By designing a search space that spans channel width, block depth, and operator type across backbone and neck, and by training 1{,}000 architectures sampled via complementary strategies on COCO-mini, we establish an initial performance database. Building on this, our \emph{Self-Evolving Predictor} iteratively enriches the database with high-value architectures, raising the ensemble LightGBM predictor's sKT from 0.694 to 0.752. Architectures discovered by predictor-guided evolutionary search surpass all official YOLOv8--YOLO12 baselines at comparable latency on COCO-mini, confirming the predictor's practical utility.
\section{Acknowledgments}
\label{sec:acknowledgement}
This work was supported by National Natural Science Foundation of China under Grant 62176007.
{
    \small
    \bibliographystyle{ieeenat_fullname}
    \bibliography{main}
}



\end{document}